\documentclass[runningheads]{llncs}

\newif\ifarxiv
\arxivtrue

\usepackage{eccv}

\usepackage{eccvabbrv}

\usepackage{graphicx}
\usepackage{booktabs}
\definecolor{cvprblue}{rgb}{0.21,0.49,0.74}
\usepackage{wrapfig}
\usepackage[table]{xcolor}
\usepackage{bm}

\ifarxiv
\else
\usepackage[accsupp]{axessibility}  
\PassOptionsToPackage{breaklinks,colorlinks,citecolor=eccvblue}{hyperref}
\fi

\ifarxiv
  \usepackage[breaklinks,colorlinks,citecolor=eccvblue]{hyperref}
\else
  \usepackage{hyperref}
  \hypersetup{linkcolor=magenta,urlcolor=magenta}
\fi

\def\ourmethod{PanoGaussian}
\definecolor{first}{rgb}{1,0.7,0.7}
\definecolor{second}{rgb}{1,0.85,0.7}
\definecolor{third}{rgb}{1,1,0.8}

\newcommand{\corrauth}{\textsuperscript{\dag}}

\begin{document}
\newlength{\figvspace}
\setlength{\figvspace}{-0.2mm}
\title{Unified Panoramic–Gaussian Representation for Monocular 4D Scene Synthesis}

\ifarxiv
\titlerunning{Unified Panoramic–Gaussian for 4D Synthesis [ECCV 2026]}
\else
\titlerunning{Unified Panoramic–Gaussian for 4D Synthesis}
\fi

\author{Yuankun Yang\inst{1} \and
Yi Wei\inst{2} \and
Wenyang Zhou\inst{2} \and
Li Zhang\inst{1}\corrauth}

\authorrunning{Y.~Yang et al.}

\institute{$^{1}$School~of~Data~Science, Fudan~University \\
$^{2}$Central Media Technology Institute, Huawei \\
\vspace{0.9em}
\href{https://github.com/LogosRoboticsGroup/PanoGaussian}{github.com/LogosRoboticsGroup/PanoGaussian}}

\maketitle
\begingroup
\renewcommand\thefootnote{}
\footnotetext{%
\corrauth\ Corresponding author:
\href{mailto:lizhangfd@fudan.edu.cn}{lizhangfd@fudan.edu.cn}%
}
\endgroup

\begin{abstract}
4D scene synthesis from monocular videos has made significant progress in recent years. However, existing methods are typically constrained by view interpolation. As a result, they struggle to infer unseen regions beyond the observed views.
In this paper, we reformulate the task as 4D scene synthesis with unseen regions, which extends beyond traditional interpolation settings. 
Camera-conditioned video generation enables unseen region synthesis by guiding generation along specified cameras.
However, these methods lack explicit 3D priors and are optimized with random camera trajectories. This design leads to severe inconsistencies under large trajectory deviations.
To address this limitation, we build a unified training and inference framework with panoramic trajectory guidance. While this design improves cross-view consistency, the panoramic representation alone fails to model dynamic content effectively. Object motion in panoramic space introduces scale and shape distortions.
To address this, we propose \ourmethod{}, a unified Panoramic-Gaussian representation that distills the panoramic representation into an explicit dynamic Gaussian representation to capture dynamic physical priors of the 4D scene. 
Experiments demonstrate that \ourmethod{} achieves consistent 4D scene synthesis even under large viewpoint variations.


\end{abstract}

\begin{figure*}[ht] 
\centering
\includegraphics[width=\textwidth]{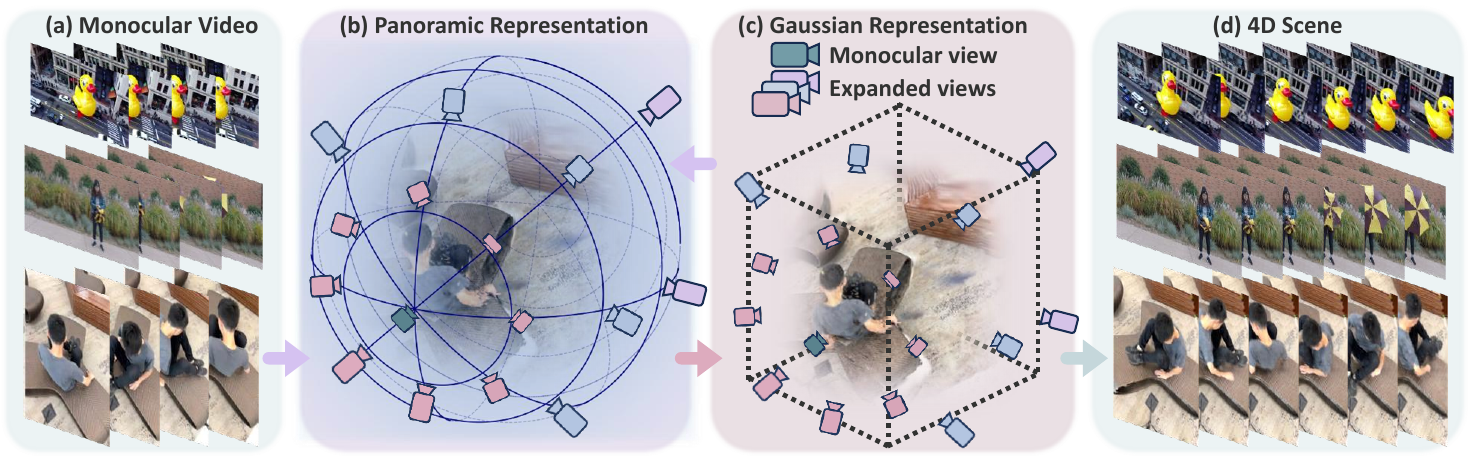}
\vspace{\figvspace}
\vspace{-5mm}
\caption{\textbf{\ourmethod{}.}
(a) Given a monocular video,
(b) \ourmethod{} constructs a panoramic representation for scene  exploration under a progressive strategy.
(c) This panoramic representation is unified with Gaussian representation to achieve geometry-preserved dynamic scene modeling.
(d) The unified Panoramic–Gaussian representation enables geometrically consistent 4D scene synthesis across unseen regions.
}
\vspace{\figvspace}
\vspace{-8mm}
\label{fig:teaser} 
\end{figure*}

\section{Introduction}
\label{sec:intro}
Reconstruction of 4D scenes from monocular videos holds considerable promise for immersive creation and virtual reality.
Recent advances mainly focus on novel view synthesis. They are typically formulated as a view interpolation task, which interpolates viewpoints within the observed camera trajectory. 
However, this task cannot be extrapolated beyond the captured range of the camera, resulting in incomplete reconstruction in unseen regions.

To address this limitation, we aim to reconstruct the entire 4D scene including both view interpolation and extension into unseen regions. 
A straightforward choice is to leverage video generation models as video generative priors to extend the unseen regions.
However, video-level generative priors generally lack 3D prior knowledge. 
This leads to distortion in 3D geometry when rendering from novel viewpoints.
Although some video generation models incorporate 3D consistency through depth re-projection or 3D datasets, they are generally optimized with small and random camera trajectories. 
As a result, these methods exhibit instability and are only effective for minor viewpoint changes.
When extrapolated to large camera angles, they produce severe geometric distortions.

Therefore, we adopt a panoramic trajectory paradigm, which aims at narrowing the gap between training and inference of generative priors to ensure geometric stability. 
This paradigm designs a fixed-angle shift strategy for geometric reprojection when training the video generative prior, enabling the model to learn consistent synthesis. During inference, we leverage the panorama coordinates to iteratively extend the scene. Panorama coordinates provide a natural global context for full-scene coverage. Therefore, it ensures spatially uniform extension of scenes while requiring minimal reliance on generative priors.
However, panoramic projections are ill-suited for modeling dynamic content. Object movements in panoramic space usually cause deformation in scale and shape. Consequently, an explicit 3D representation is still necessary to preserve geometry while capturing dynamics.  

To this end, we propose \ourmethod{}, which unifies the panoramic representation with dynamic Gaussian Splatting representation for geometry-preserved 4D scene modeling, as illustrated in~\cref{fig:teaser}. Specifically, \ourmethod{} first leverages the panorama coordinates to design the camera trajectory for iterative view expansion. These expanded camera trajectories are processed by the video generative prior to synthesizing novel views. The generated results are subsequently distilled into the dynamic Gaussian Splatting representation for geometrically consistent 4D reconstruction.

Our main contributions are summarized as follows:
1) We propose \ourmethod{}, a unified Panoramic-Gaussian representation that distills video generative priors into explicit 3D dynamic Gaussians through masked refinement, preserving geometry and motion under large viewpoint changes.
2) We design progressive expansion with fixed-geometry warping and region-wise supervision that structurally prevents error accumulation during unseen-region synthesis.
3) \ourmethod{} achieves consistent 4D scene synthesis, maintaining geometric fidelity even under large viewpoint changes.

\section{Related work}
\label{sec:related}

\subsubsection{Dynamic Novel View Synthesis} is commonly formulated as view interpolation from a set of posed images.
Early dynamic reconstruction approaches explored neural scene representations (NeRFs) by 
treating time as an additional coordinate~\cite{Ramasinghe2024blirf, Wang2021NeuralTF, song2023nerfplayer, li2020neuralsceneflow, Gao2021DynNeRF, gao2022dycheck, Fridovich2023kplanes}. 
More recent efforts extended Gaussian Splatting (3DGS) to the 4D domain by modeling 3D Gaussian trajectories over time~\cite{luiten2023dynamic, sun20243dgstream, Duisterhof2023MDSplattingLM, das2023neuralparametric, Katsumata2023AnE3, lin2023gaussianflow, li2023spacetimetrajs, yu2023cogs}, learning time-conditioned deformation networks~\cite{wu20234dgaussians, yang2023deformable3dgs, Liang2023GauFReGD, Guo2024Motionaware3G}, or directly parameterizing Gaussians with 4D means and covariances~\cite{Duan20244DGS, yang2023gs4d}.
In monocular settings~\cite{zhou2025feature4x,wang2025gflow,chu2024dreamscene4d,li2024self}, Gaussian Marbles~\cite{stearns2024dynamic} introduced isotropic Gaussian constraints to jointly model motion and appearance. MoSca~\cite{lei2024mosca} integrated 2D foundation model priors to guide motion scaffold optimization. USplat4D~\cite{guo2025uncertainty} modeled motion uncertainty to improve rendering robustness. Shape of Motion~\cite{wang2024shape} exploited the low-dimensional structure of scene motion to regularize 3D trajectories.
Recent feed-forward approaches~\cite{liang2024feed, shen2025seeing, lin2025movies,xu20254dgt,lin2025dgs, luo2025instant4d, wu20254d} aimed to directly estimate 3DGS attributes to eliminate per-scene optimization.
Despite these advances, all existing methods struggle to extrapolate beyond the observed camera range, yielding incomplete reconstructions in unseen regions.

\subsubsection{Camera-Conditioned Video Generation}
aims to synthesize controllable videos along specified camera trajectories. Early approaches~\cite{hong2022cogvideo,blattmann2023stable} pioneered conditional video generation by conditioning diffusion architecture~\cite{ho2020denoising,nichol2021improved,rombach2022high,zhang2023adding} on the first-frame latent representation for temporal coherence.
Building upon these foundation models, recent camera-conditioned video generation models~\cite{van2024generative, ren2025gen3c,you2024nvs, yu2025trajectorycrafter,yesiltepe2025dynamic, wu2025video, chen2025reconstruct,cao2025uni3c,zheng2026versecrafter} 
condition video diffusion on projected 3D point clouds derived from target camera trajectories for geometry-aware synthesis.
In contrast, another related line of methods~\cite{he2024cameractrl, he2025cameractrl,bai2025recammaster,zhou2025stable, luo2025camclonemaster} discards explicit reprojection, instead encoding camera motion through learned or Plücker-based camera embeddings to facilitate flexible viewpoint control.
However, these frameworks lack holistic 3D priors, often leading to spatial inconsistencies under large viewpoint changes.
\ourmethod{} addresses this limitation by injecting a unified Panoramic-Gaussian representation that enables complete 4D scene perception and enforces geometric consistency throughout the generated videos.

\subsubsection{Panoramic Scene Synthesis} captures a wide, continuous field of view for immersive scene representation.
Traditional methods design panoramas as concentric mosaics~\cite{shum1999rendering}, layered depth map~\cite{shade1998layered,zheng2007layered} or mesh representations~\cite{hedman2018instant,hedman2017casual} for improved rendering of object surfaces.
Recent panoramic scene reconstruction approaches optimize neural radiance fields~\cite{wang2024perf, huang2022360roam,gu2022omni,lu2024pano, chugunov2024neural} or 3D Gaussian Splatting~\cite{bai2025360,lee2024odgs,chen2025splatter,li2025omnigs} directly from a single panorama to render static volume of the whole scene.
These methods further advance the generative panorama outpainting~\cite{zhang2023diffcollage,wu2023panodiffusion,ai2024dream360,lu2024autoregressive,wang2024360dvd,li20244k4dgen,liu2025dynamicscaler} through the diffusion denoising process.
However, directly extending panorama to 4D scene reconstruction from monocular videos remains challenging, as object motion in panoramic space often causes geometric distortion. To address this, \ourmethod{} introduces a Panoramic-Gaussian representation that preserves geometry while capturing dynamic object motion.

\section{Preliminaries}
\label{sec:preliminary}

\subsubsection{3D Gaussian Splatting} explicitly represents a scene as a collection of 3D Gaussians~\cite{kerbl20233Dgaussians}.
Each Gaussian is defined by color $\mathbf{c} \in \mathbb{R}^3$, opacity $\alpha \in \mathbb{R}$, position mean $\mathbf{\mu} \in \mathbb{R}^3$ and covariance matrix $\Sigma \in \mathbb{R}^{3\times 3}$.
However, Gaussian Splatting typically relies on pixel-level supervision during optimization, which restricts its ability to generalize beyond observed views.

\subsubsection{Video Generative Priors} enable controllable view synthesis by guiding video generation along a specified camera trajectory.
Given an input video  $\bm{x}_0 \in \mathbb{R}^{\text{n}\times \text{3}\times \text{h}\times \text{w}}$, the forward diffusion process progressively injects Gaussian noise as $\bm{x}_t = \alpha_t\bm{x}_0 + \sigma_t \epsilon$,
where $\alpha_t \in (0,1)$ and $\sigma_t$ are the signal and noise scaling coefficients in timestep $t$.
The reverse process learns to recover the clean video distribution by progressively denoising:
\begin{equation}
\min_{\theta} \mathbb{E}_{t\sim\mathcal{U}(0,1),\epsilon\sim\mathcal{N}(\bm{0},\bm{I})}[\|\epsilon_\theta(\bm{x}_t,t) - \epsilon\|_2^2],
\end{equation}
where $\bm{\epsilon}_\theta(\bm{x}_t,t)$ is the network-predicted noise parameterized by $\theta$, allowing progressive recovery of $\bm{x}_0$ from $\bm{x}_t$.
Despite their strong generative ability, video diffusion usually lacks explicit 3D geometric understanding.
They are effective for small viewpoint changes, but often fail under large camera motions.

\subsubsection{Panoramic Representation} captures a complete $360^\circ \times 180^\circ$ view of a scene. It is usually defined in a spherical domain.
This representation is parameterized by the coordinates $(\phi, \theta)$, where $\phi$ is the azimuth and $\theta$ the elevation.
Although providing global scene coverage, the panoramic projection introduces geometric distortions. Object motion in panoramic space leads to scale and shape deformation.
Therefore, bridging this gap between panoramic representation and 3D geometry is essential for geometry-aware 4D scene synthesis.
\begin{figure*}[t] 
\centering
\includegraphics[width=\textwidth]{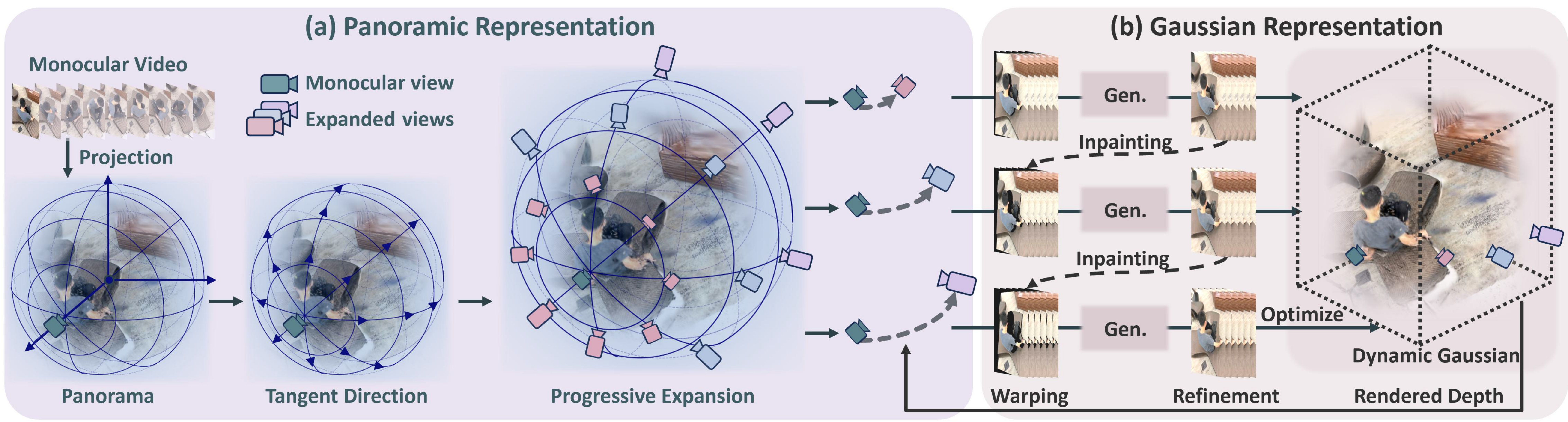}
\vspace{\figvspace}
\caption{
\textbf{Architecture Overview.}
(a) \ourmethod{} begins by reprojecting the monocular video into panoramic coordinates for full-scene exploration (\cref{sec:panorama}).
Under this panorama, tangent directions are constructed to progressively guide trajectory expansion for unseen regions (\cref{sec:expansion}).
(b) \ourmethod{} employs Gaussian-rendered depth to warp the original video frames along these expanded trajectories.
The warped videos are then inpainted and refined by a video generative prior pretrained under panoramic trajectory paradigm.
The refined results are distilled into Dynamic Gaussians for geometry-preserving 4D scene synthesis (\cref{sec:gaussian}). 
}
\vspace{\figvspace}
\label{fig:pipeline} 
\end{figure*}

\section{Methods}
\label{sec:methods}

Given a monocular video, \ourmethod{} aims to extend conventional view interpolation to geometry-aware 4D scene reconstruction with coverage of unseen regions.
Our core contribution is a unified Panoramic-Gaussian representation that distills panoramic trajectory-guided generative outputs into explicit 3D dynamic Gaussians, as illustrated in~\cref{fig:pipeline}.
To achieve this, we adopt a panoramic trajectory paradigm for camera-conditioned video priors (\cref{sec:panorama}), progressively expand scene coverage under fixed-geometry warping (\cref{sec:expansion}), and unify panoramic exploration with dynamic Gaussian optimization via masked refinement (\cref{sec:gaussian}).

\subsection{Panoramic Trajectory Paradigm}
\label{sec:panorama}
To condition video generative priors on target viewpoints, we project observations into a panoramic domain and keep training and inference under the same warping protocol.
Specifically, we apply a fixed panoramic trajectory protocol for both training and inference of the video generative priors.

\subsubsection{Panoramic Trajectories Training.}
We design a fixed panoramic angle to guide the model in learning panoramic-aware motion and appearance generation during training.
Specifically, for each source video $V_{src,tr}$ in the training set, we construct a panoramic trajectory following a random direction $d_{tr}$ and a fixed panoramic elevation $\theta$.
This panoramic trajectory is generated by uniform sampling along the panoramic direction to simulate the camera motion.
Following the exact configuration of previous projection-based generative priors~\cite{chen2025reconstruct}, we conduct a double reprojection strategy along the defined trajectory to produce a warped covisible video $V_{cov,tr}$.
The masked video is then applied to train the video diffusion prior.
It ensures that the model learns to synthesize content consistent with panoramic viewpoints.
This process enables video generative priors to internalize panoramic spatial continuity.

\subsubsection{Panoramic Paradigm Inference.}
In the inference stage, we design a progressive extension of the 4D scene within the panoramic coordinate system.
Specifically, for each video frame $I_t \in V_{\text{src},\text{inf}}$, we extract its corresponding depth map $D_t$ and camera poses (rotation $R_t$, intrinsics $K_t$) from the optimized 4D Gaussian Splatting (4D GS) backbone.
We unproject these pixels into a sequence of 3D point clouds ${P}_t=\Phi^{-1}([I_{t},D_{t}], R_{t},K_{t})$ through inverse perspective projection.
Crucially, we assume that depth $D_t$ and poses $[R_t, K_t]$ derived from 4D GS are fixed during this step. 
Because this process inherently operates in 3D world-space coordinates, camera translation, dynamic motion, and parallax are naturally handled without requiring any image-space approximations.
The center of the projected point cloud is defined as the origin of the panorama $P^{c}=(x^c,y^c,z^c)$.
We then map the 3D points ${P}_t$ into the panoramic space. The full projection implementation details are provided in the supplementary material.
This ensures that the first camera direction corresponds to $x$-axis for uniform scene exploration.

In this way, we achieve both training and inference in correspondence under the panoramic paradigm. 
This alignment reduces domain gaps between optimization and generation of the generative prior.
We then perform progressive view expansion (\cref{sec:expansion}) within this aligned system.

\subsection{Progressive Expansion Strategy}
\label{sec:expansion}
After mapping all video pixels into the panoramic coordinate system, \ourmethod{} applies a progressive expansion strategy to gradually explore the full 4D scene.
Naive or random view expansion often suffers from error accumulation and inconsistency between multiple trajectories. 
To address this inefficiency, we design a regularized expansion method that ensures comprehensive yet coherent scene exploration.

\subsubsection{Panoramic Trajectory Construction.}
For view expansion, we use $M$ evenly spaced directions on the panoramic sphere:
\begin{equation}
d_m = (\cos(\tfrac{2\pi m}{M}), \sin(\tfrac{2\pi m}{M})),
\quad m = 0,\dots,M-1,
\label{eq:dir1}
\end{equation}
where each $d_m$ represents a tangent direction for view expansion.
Along each direction $d_m$, the camera view is expanded $E$ times with an angular step $\alpha$.
The accumulated rotation in the expansion step $e$ is expressed as:
\begin{equation} \mathbf{s}_{m,e} = \mathbf{n}\cos(e\alpha) + (\mathbf{a}_m \!\times\! \mathbf{n})\sin(e\alpha), \quad e=0,\dots,E, 
\label{eq:dir2}
\end{equation}
where $\mathbf{n}$ is the initial camera viewing vector, and $\mathbf{a}_m$ denotes the unit rotation axis orthogonal to $\mathbf{n}$ in the tangent plane defined by $d_m$.
This procedure produces $E$ uniformly interpolated panoramic trajectories that progressively cover the entire scene.
Each trajectory contains $T$ frames over time, denoted as $P_{e,t}$ where $e = 0,\dots,E-1$ and $t=0,\dots,T-1$.

\subsubsection{Progressive Scene Expansion.}
Following prior camera-controlled generation pipelines~\cite{chen2025reconstruct,yu2025trajectorycrafter}, we warp source frames onto each trajectory and inpaint masked regions with the video diffusion prior.
For each target video frame $I_{t} \in V_{\text{src},\text{inf}}$, we first warp it onto the panoramic trajectory $P_{e,t}$, resulting in the warped $I_{e,t}$.
These warped frames $\{{I_{e,t}}\}_{t=0}^{T-1}$ form the intermediate video $V_{e, \text{warped}}$.
We then apply progressive generation with the video diffusion prior to the panoramic representation.
Specifically, the warped intermediate video $V_{e, \text{warped}}$ and a binary mask indicating the unobserved regions are directly provided as conditioning inputs for the video diffusion to perform inpainting.
At every step, warping uses the original observed video $V_{\text{src},\text{inf}}$ with fixed 4D GS geometry rather than previously hallucinated outputs, and the mask confines supervision to newly generated regions, structurally preventing error accumulation across steps.
Each warped video $V_{e, \text{warped}}$ is successively refined to produce $V_{e,\text{refined}}$. This procedure extends the view and recovers details in unobserved regions.
After each refinement step, the updated video $V_{e,\text{refined}}$ is warped to the next unrefined trajectory $V_{e+1, \text{warped}}$ to fill the missing content, while preserving areas that have already warped.
This iterative process continues until all panoramic trajectories are refined, resulting in videos that are spatially continuous and mutually consistent across all directions.
The refined videos are then used to supervise the Panoramic-Gaussian representation in~\cref{sec:gaussian}.

\subsection{Panoramic-Gaussian Representation}
\label{sec:gaussian}

The panoramic coordinates provide a natural global context for capturing the entire scene. However, object motion in panoramic space often leads to deformation in scale and shape. This deformation disrupts the underlying physical dynamics, resulting in inconsistent geometry and unrealistic motion.
To address this, \ourmethod{} introduces a composite Panoramic-Gaussian representation. This design leverages panoramic coverage to provide global consistency while applying Gaussian priors to preserve geometry.

\subsubsection{Unified Gaussian Geometry.}
We first optimize dynamic Gaussian Splatting on reference videos to obtain a stable Gaussian representation of the observed scene. From this representation, we extract panoramic trajectories $P_{e,t}$ as described in~\cref{sec:expansion}, and render  $V_{e,\text{render}}$ based on these camera trajectories from the Gaussian representation. 
Previous camera-controlled video generation methods~\cite{yu2025trajectorycrafter,ren2025gen3c} estimate depth independently for each frame. This operation introduces large depth estimation errors, causing severe inconsistency and broken geometry.
In contrast, our unified Panoramic-Gaussian design directly projects the stable depth maps from known views to novel viewpoints as geometric references. Unseen regions are seamlessly inpainted by the generative model without requiring pre-existing depth, and their newly hallucinated geometry is subsequently distilled back to update the 4D Gaussian model.
This progressive depth initialization ensures consistent geometric alignment and prevents structure collisions during panoramic rendering.

\subsubsection{Refinement Regulation.}
After obtaining the refined video $V_{e,\text{refined}}$, we introduce a refinement regulation to enforce pixel-wise consistency.
Specifically, we design a masked mean squared error (MSE) between Gaussian rendering and panoramic-trajectory refined videos:
\begin{equation}
L_{\text{refine}} = \text{MSE}(M_{e} \cdot V_{e,\text{refined}}, M_{e} \cdot V_{e,\text{render}}).
\label{eq:refine}
\end{equation}
Here, $M_e$ is a binary mask indicating the unknown regions in the warped video $V_{e,\text{warped}}$. It is generated by identifying pixels with no projected points from the known views. To ensure consistency across the refinement process, $M_e$ is kept fixed during the current expansion step $e$, focusing the loss strictly on the newly hallucinated content.
This refinement is jointly optimized with the source video $V_{\text{src},\text{inf}}$, guided by all geometric regularization in MoSca~\cite{lei2024mosca}. 
We update the progressive scene expansion after specific Gaussian optimization iterations. 
Specifically, new Gaussians are initialized from the inpainted regions and optimized via the MSE loss against the refined video, without requiring explicit depth estimates from the diffusion model. This could produce improved Gaussian geometry for panoramic warping and generate updated panoramic trajectories to supervise the dynamic Gaussian.

Through this unified optimization, each video pixel along the panoramic trajectory directly corresponds to its Gaussian-rendered counterpart. This correspondence forms a pixel-to-pixel alignment between panoramic and Gaussian representations. As a result, \ourmethod{} effectively distills panoramic exploration into Gaussian representation, preserving both global panorama exploration and the physical priors of Gaussian.

\begin{figure*}[t] 
\centering
\includegraphics[width=\linewidth]{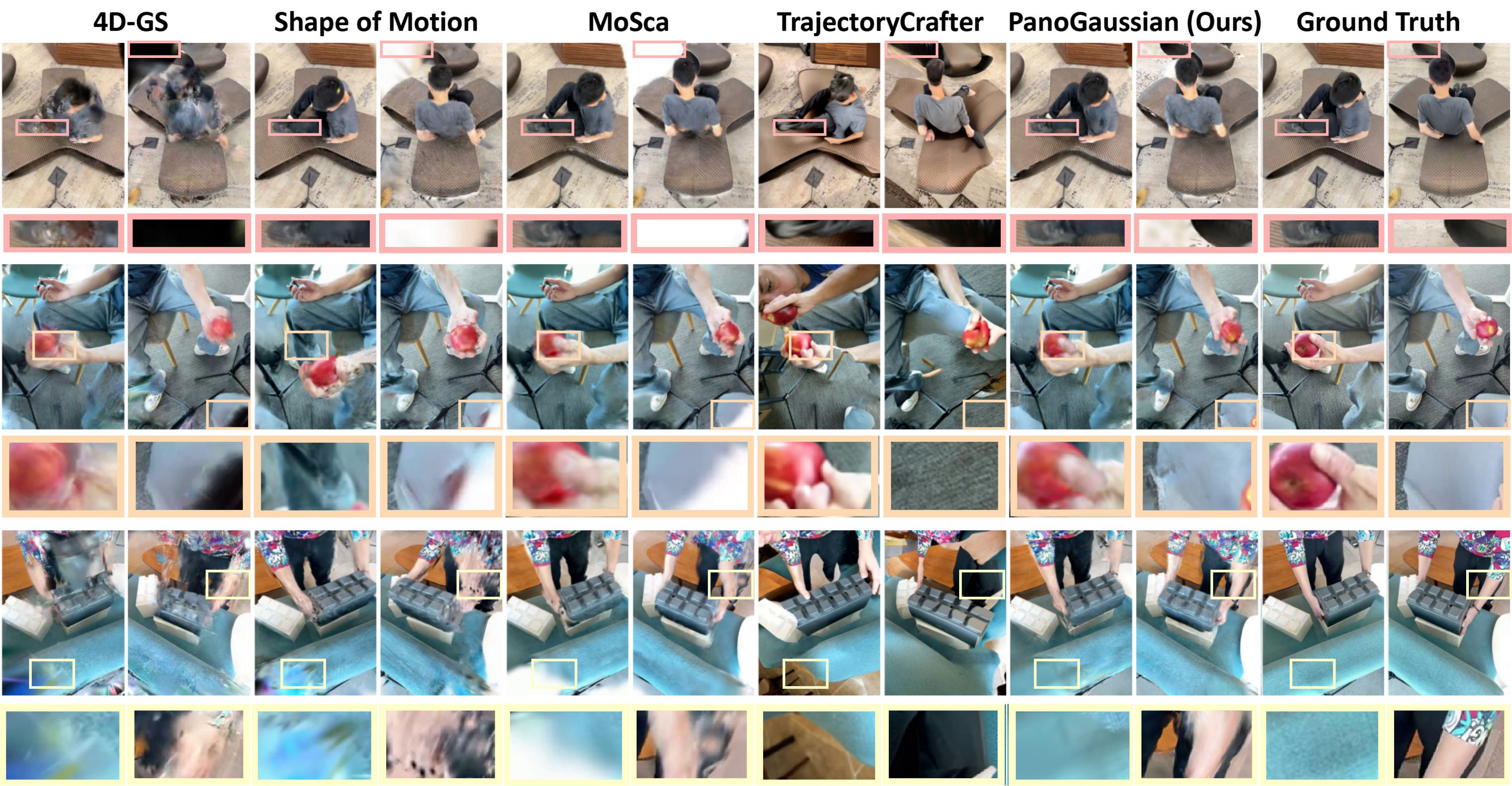}
\vspace{\figvspace}
\caption{
\textbf{Qualitative comparison of \ourmethod{} on DyCheck~\cite{li2023dynibar}. }
The red, orange, and yellow boxes highlight detailed synthesis regions in each case.
Previous state-of-the-art reconstruction-based approaches such as 4D Gaussian Splatting~\cite{wu20244d}, Shape-of-Motion~\cite{wang2024shape} and MoSca~\cite{lei2024mosca} struggle to infer unseen regions. These methods lead to missing (white) areas  and blurred reconstructions in boxed areas. 
Although the generative method TrajectoryCrafter~\cite{yu2025trajectorycrafter} can infer unseen regions, it often produces geometrically inconsistent results.
In comparison, \ourmethod{} achieves both geometry-consistent and complete 4D scene synthesis.
}
\vspace{\figvspace}
\label{fig:results} 
\end{figure*}

\begin{table*}[ht]
    \centering
    \caption{\textbf{Comparison with reconstruction-based and generation-based methods on the DyCheck iPhone dataset ~\cite{li2023dynibar}.}
    ``G.S.", ``Diff", ``Feed" means Gaussian Splatting methods, diffusion-based approaches, feed-forward methods, respectively.
    Best, second-best, and third-best results are highlighted in \colorbox{first}{red}, \colorbox{second}{orange}, and \colorbox{third}{yellow}, respectively.
    Our model consistently achieves the best performance in all metrics.
    }
    \vspace{\figvspace}
    \label{tab:performance_iphone}
     \resizebox{\columnwidth}{!}{
    \begin{tabular}{lcccccccccc}
    \toprule
    Method & Type & $\uparrow$ PSNR & $\uparrow$ SSIM & $\downarrow$ LPIPS & $\uparrow$ mPSNR & $\uparrow$ mSSIM & $\downarrow$ mLPIPS & $\uparrow$ uPSNR & $\uparrow$ uSSIM & $\downarrow$ uLPIPS \\
    \midrule
    4D G.S.~\cite{wu20244d} & G.S. & 15.71 & 0.450 & 0.398 & 16.54 & 0.594 & 0.347 & 12.82 & 0.302 & 0.452 \\
    Shape-of-Motion~\cite{wang2024shape} & G.S. & \cellcolor{third}16.79 & \cellcolor{third}0.510 & \cellcolor{third}0.391 & 17.32 & 0.598 & \cellcolor{third}0.296 & \cellcolor{third}{{14.84}} &  0.321 &  0.448 \\
    MoSca~\cite{lei2024mosca} & G.S. & \cellcolor{second}{\underline{17.31}} & \cellcolor{second}{\underline{0.573}} & \cellcolor{second}{\underline{0.354}} & \cellcolor{second}{\underline{19.32}} & \cellcolor{second}{\underline{0.706}} & \cellcolor{second}{\underline{0.264}} & 14.23 & 0.295 & 0.461 \\
    BulletGen~\cite{rozumnyi2025bulletgen} & G.S. & -- & -- & -- & 16.78 & \cellcolor{third}0.640 & 0.390 & -- & -- & -- \\
    Vivid4D~\cite{huang2025vivid4d} & G.S. & -- & -- & -- & 15.20 & 0.500 & 0.493 & -- & -- & -- \\
    \midrule
    Cat4D~\cite{wu2024cat4d} & Diff. & 15.91 & 0.427 & 0.398 & 17.39 &  0.607 & 0.341 & 14.06 & \cellcolor{third}{{0.351}} & \cellcolor{third}{{0.428}} \\
    TrajectoryCrafter~\cite{yu2025trajectorycrafter} & Diff. & 13.58 & 0.373 & 0.483 & 15.48 & 0.523 & 0.426 & 10.88 & 0.315 & 0.441 \\
    CogNVS~\cite{chen2025reconstruct} & Diff. & 16.94 & 0.449 & 0.598 & \cellcolor{third}18.63 & 0.614 & 0.290 & \cellcolor{second}{\underline{14.97}} & \cellcolor{second}{\underline{0.371}} & \cellcolor{second}{\underline{0.487}} \\
    \midrule
    MoVieS~\cite{lin2025movies} & Feed & -- & -- & -- & 18.46 & 0.589 & 0.309 & -- & -- & -- \\
    BTimer~\cite{liang2024feed} & Feed & -- & -- & -- & 16.52 & 0.570 & 0.338 & -- & -- & -- \\
    4DGT~\cite{xu20254dgt} & Feed & 15.04 & 0.446 & 0.423 & 16.12 & 0.556 & 0.408 & 12.11 & 0.301 & 0.450 \\
    4D-Fly~\cite{wu20254d} & Feed & -- & -- & -- & 17.03 & 0.60 & 0.37 & -- & -- & -- \\   
    \midrule
    \ourmethod{} (Ours) & Pano & \cellcolor{first}{\textbf{18.71}} & \cellcolor{first}{\textbf{0.598}} & \cellcolor{first}{\textbf{0.323}} & \cellcolor{first}{\textbf{19.85}} & \cellcolor{first}{\textbf{0.741}} & \cellcolor{first}{\textbf{0.239}} & \cellcolor{first}{\textbf{16.72}} & \cellcolor{first}{\textbf{0.495}} & \cellcolor{first}{\textbf{0.385}} \\
    \bottomrule
    \end{tabular}
    }
    \vspace{\figvspace}
    \end{table*}

\section{Experiments}
\label{sec:experiments}

\subsection{Experimental Settings}

\subsubsection{Datasets.}
We adopt two standard benchmarks: the DyCheck IPhone dataset~\cite{li2023dynibar} and the Nvidia Dynamic dataset~\cite{yoon2020novel}. DyCheck contains long monocular video sequences with significant camera motion and challenging unseen regions in the testing split (i.e., test camera trajectories deviate significantly from the training views), allowing us to assess the exploration capabilities of \ourmethod{}.
We further demonstrate the generalization of \ourmethod{} on in-the-wild video sequences to verify robustness under unconstrained settings.

\subsubsection{Metrics.}
We evaluate reconstruction quality using standard metrics: PSNR, SSIM, and LPIPS.
Following DyCheck, we also report metrics (mPSNR, mSSIM, mLPIPS) on covisible regions.
Since our task extends view interpolation to full 4D scene synthesis with unseen regions, we introduce additional metrics (uPSNR, uSSIM, uLPIPS) to measure performance exclusively on purely unseen areas. These unseen metrics are computed on the inverse of the DyCheck masks, specifically isolating regions outside the training camera covisibility.

\subsubsection{Baselines.}
We compare our method with a broad range of monocular 4D reconstruction approaches.
NeRF-based methods~\cite{li2021neural,pumarola2021d,pumarola2021dnerf,tretschk2021non,fang2022fast,park2021nerfies,park2021hypernerf,gao2021dynamic,tian2023mononerf,miao2024ctnerf} extend NeRFs to model temporal dynamics.
Gaussian Splatting methods~\cite{wu20244d,wang2024shape,zhou2024dynpoint,lei2024mosca,rozumnyi2025bulletgen,guo2025uncertainty} achieve efficient spatio-temporal reconstruction through dynamic Gaussian primitives.
Diffusion-based approaches~\cite{wu2024cat4d,yu2025trajectorycrafter,chen2025reconstruct} leverage camera-controlled video diffusion models for novel-view generation.
We also compare to feed-forward methods~\cite{lin2025movies,liang2024feed,xu20254dgt} that directly estimate Gaussian Splatting attributes without per-scene optimization.
Many of these approaches are concurrent with our work~\cite{chen2025reconstruct,lin2025movies,liang2024feed,xu20254dgt,rozumnyi2025bulletgen,guo2025uncertainty}, exist only as preprints, or have not made their code publicly available. We report on their results as provided in their respective papers. Note that recent camera-conditioned models (e.g., Gen3C~\cite{ren2025gen3c}, Recammaster~\cite{bai2025recammaster}) are not designed for reconstruction, thus are excluded. In our tables, '-' indicates unavailable data due to unreleased full code. 

\subsubsection{Implementation details.}
We implement our panoramic trajectory training based on the pretrained TrajectoryCrafter~\cite{yu2025trajectorycrafter}.
Training is conducted on the large-scale OpenVid dataset~\cite{nan2024openvid}, which provides diverse dynamic scenes and high-fidelity visual details suitable for optimizing generative priors.
The video frames are preprocessed to a resolution of $384\times672$ and a length of 49 frames.
To avoid overfitting, only the cross-attention and patch embedding layers in the Ref-DiT blocks are optimized.
The whole training takes 25{,}000 iterations with a learning rate of $2\times10^{-6}$.
For Panoramic Paradigm Inference, we adopt a spherical coordinate system with azimuthal angle $\phi \in [-\pi, \pi)$ and polar angle $\theta \in [-\frac{\pi}{2}, \frac{\pi}{2}]$.
The panorama resolution is set to $w_p = 2048$ and $h_p = 2048$.
We use $M=8$ directional views for full-scene coverage, $E=6$ expansion steps, and an angular step of $\alpha = 15^\circ$ in the progressive expansion strategy.
We employ MoSca~\cite{lei2024mosca} as the dynamic Gaussian backbone for our method and all Gaussian-based comparisons to ensure fair evaluation.
The refinement weight is set to $\lambda_{\text{refine}} = 0.1$.
Gaussian optimization runs for 10{,}000 iterations. Specifically, every 2,000 iterations of GS optimization, we perform progressive panoramic expansion via the video diffusion prior.
Total per-scene optimization time is approximately 2.0h, rendering at 32.42FPS.
Our 2.0h runtime consists of 1.4h for 4D Gaussian Splatting fitting and only 0.6h for progressive diffusion generation. The generation time scales linearly with the number of expansion steps $E$ and tangent directions $M$.
Compared to baselines, this is significantly faster than Shape of Motion~\cite{wang2024shape} (4.2h) and comparable to MoSca~\cite{lei2024mosca} (1.3h).
For evaluation on DyCheck and Nvidia datasets, we report quantitative results using Gaussian representation renderings.
For extreme view exploration, we present the refined final-step results in the panoramic representation to emphasize visual quality.

\subsection{Novel View Synthesis}
\subsubsection{Results on DyCheck~\cite{li2023dynibar}.}
As shown in~\cref{tab:performance_iphone} and~\cref{fig:results}, \ourmethod{} consistently achieves the best performance in all metrics.
NeRF-based and Gaussian Splatting methods primarily focus on view interpolation within the observed camera trajectories.
As a result, they cannot infer unseen regions, leading to missing reconstructions in the boxed areas of~\cref{fig:results}. 
This limitation also causes a notable performance drop when switching from mPSNR (covisible regions) to PSNR (full scene) in~\cref{tab:performance_iphone}.
Video diffusion approaches, on the other hand, often produce geometrically distorted results due to the absence of explicit 3D priors.
In comparison, \ourmethod{} leverages a unified Panoramic-Gaussian representation to achieve geometry-consistent 4D scene synthesis, demonstrating superior structural fidelity even under viewpoint changes.

\subsubsection{Results on Nvidia~\cite{yoon2020novel}.}
\begin{figure}[t] 
\centering
\includegraphics[width=\columnwidth]{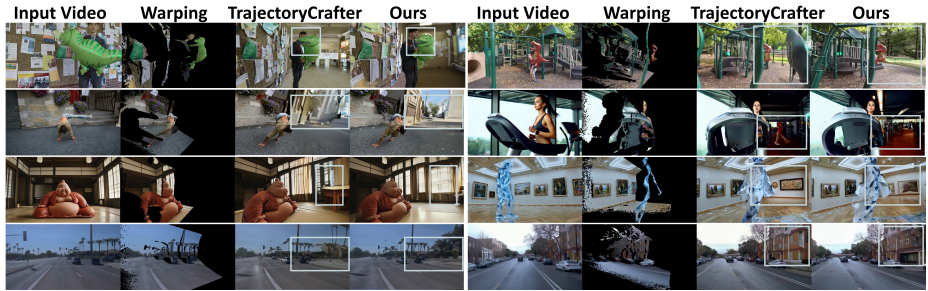}
\vspace{\figvspace}
\caption{
\textbf{Scene exploration under extreme camera transitions.}
We visualize warping results under camera trajectories exhibiting drastic viewpoint changes beyond the original range on Nvidia~\cite{yoon2020novel} and in-the-wild dynamic videos.
Previous method~\cite{yu2025trajectorycrafter} produces severe geometric distortions and artifacts in the green boxed areas. In comparison, \ourmethod{} preserves geometry and synthesizes consistent and realistic scenes even under extreme camera transitions.
}
\vspace{\figvspace}
\label{fig:exploration} 
\end{figure}

Unlike DyCheck, the Nvidia testing split consists of scenes with minimal viewpoint variation and no truly unseen regions.
Consequently, \ourmethod{} achieves performance comparable to existing methods, as shown in~\cref{fig:exploration} and~\cref{tab:nvidia}.
The extrapolation gains observed on DyCheck do not fully manifest here.
This result is expected, since our approach is primarily designed to explore and reconstruct unseen regions, while also preserving interpolation quality within observed viewpoints.
To better highlight this strength, we additionally perform an extreme view exploration experiment on the Nvidia dataset without ground-truth supervision, where \ourmethod{} qualitatively synthesizes coherent and geometrically consistent novel views.
\setlength{\tabcolsep}{10pt} 
\begin{table*}[t]
\centering
\caption{\textbf{Quantitative comparison with reconstruction-based methods on Nvidia~\cite{li2023dynibar}.}
``NeRF", ``G.S.", ``Feed" means NeRF-based, Gaussian Splatting, and feed-forward methods, respectively.
Best, second-best, and third-best results are highlighted in \colorbox{first}{red}, \colorbox{second}{orange}, and \colorbox{third}{yellow}, respectively.}
\label{tab:nvidia}
\vspace{\figvspace}
\setlength{\tabcolsep}{0pt} 
\begin{tabular*}{\linewidth}{@{\extracolsep{\fill}} l ccc p{0.2em} l ccc} 
\toprule
Method & Type & $\uparrow$ PSNR & $\downarrow$ LPIPS & & Method & Type & $\uparrow$ PSNR & $\downarrow$ LPIPS \\
\cmidrule{1-4} \cmidrule{6-9}
NSFF~\cite{li2021neural} & NeRF & 24.33 & 0.199 & & 4D G.S.~\cite{wu20244d} & G.S. & 21.45 & 0.199 \\
D-NeRF~\cite{pumarola2021d} & NeRF & 21.49 & 0.232 & & GaussianMarbles~\cite{stearns2024dynamic} & G.S. & 22.32 & 0.129 \\
DynNeRF~\cite{gao2021dynamic} & NeRF & 26.10 & 0.082 & & MoSca~\cite{lei2024mosca} & G.S. & \cellcolor{second}{\underline{26.72}} & \cellcolor{third}{0.070} \\
MonoNeRF~\cite{tian2023mononerf} & NeRF & 25.62 & 0.106 & & BulletGen~\cite{rozumnyi2025bulletgen} & G.S. & 17.02 & 0.386 \\
CTNeRF~\cite{miao2024ctnerf} & NeRF & 26.13 & 0.082 & & MoVieS~\cite{lin2025movies} & Feed & 19.16 & 0.315 \\
HyperNeRF~\cite{park2021hypernerf} & NeRF & 17.60 & 0.367 & & 4D-Fly~\cite{wu20254d} & Feed & 22.52 & 0.14 \\
DynPoint~\cite{zhou2024dynpoint} & NeRF & \cellcolor{third}26.53 & \cellcolor{first}{\textbf{0.068}} & & \ourmethod{} (Ours) & Pano & \cellcolor{first}{\textbf{26.75}} & \cellcolor{second}{\underline{0.069}} \\
\bottomrule
\end{tabular*} 
\vspace{\figvspace}
\end{table*}

\subsubsection{Qualitative Results on Kubric-4D~\cite{vanhoorick2024gcd}.}
\begin{figure}[t]
\centering
\includegraphics[width=\linewidth]{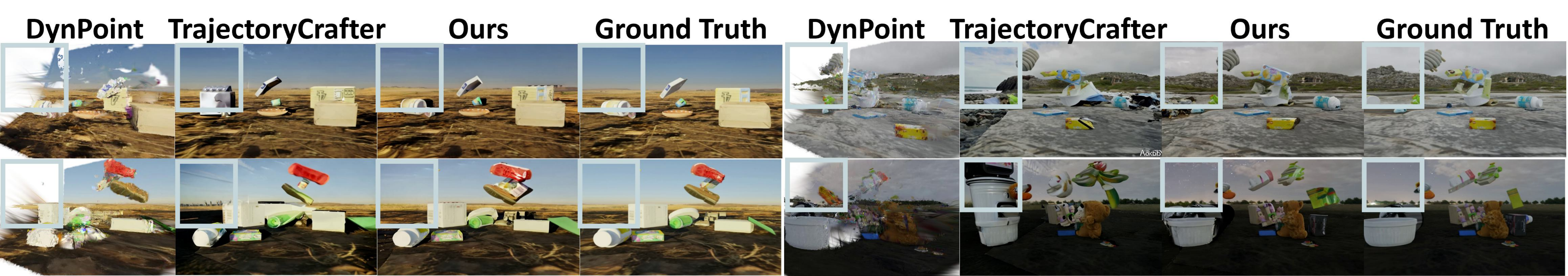}
\vspace{\figvspace}
\caption{\textbf{Qualitative comparison on Kubric-4D~\cite{vanhoorick2024gcd}.}
Under large camera displacement, baselines fail to synthesize coherent unseen regions, while \ourmethod{} preserves consistent structure and appearance.}
\label{fig:kubric}
\vspace{\figvspace}
\end{figure}

Unlike the Nvidia dataset, Kubric-4D features substantial camera motion with genuinely unseen regions, making it a more suitable benchmark for evaluating extrapolation.
As shown in~\cref{fig:kubric}, reconstruction-based and diffusion-based baselines produce incoherent or missing content in unseen regions, whereas \ourmethod{} maintains structurally and visually consistent synthesis.

\subsubsection{Extreme View Exploration.}
To evaluate the generalizability of \ourmethod{}, we test its robustness under extreme camera transitions far beyond the original trajectory. 
We conduct scene exploration on in-the-wild dynamic videos containing multiple objects and complex motions.
Since extreme in-the-wild scene exploration lacks ground truth, quantitative evaluation is infeasible. Furthermore, we focus our qualitative comparison on the most relevant state-of-the-art TrajectoryCrafter~\cite{yu2025trajectorycrafter}.
NeRF-based and Gaussian-based reconstruction methods fail to infer unseen regions in this challenging situation without supervision. 
As illustrated in~\cref{fig:exploration}, \ourmethod{} achieves consistent 4D scene synthesis under a drastic viewpoint transition.
In comparison, previous video diffusion methods~\cite{yu2025trajectorycrafter} suffer from severe geometric distortion, such as the deformed balloon in the first row, physically inconsistent buildings in the second and third rows, and the distorted human in the last row.
These results demonstrate that our Panoramic-Gaussian framework effectively unifies the strengths of geometric reconstruction and video generative priors for robust 4D scene synthesis, enabling physically consistent 4D scene synthesis.

\begin{figure}[t]
\centering
\includegraphics[width=\linewidth]{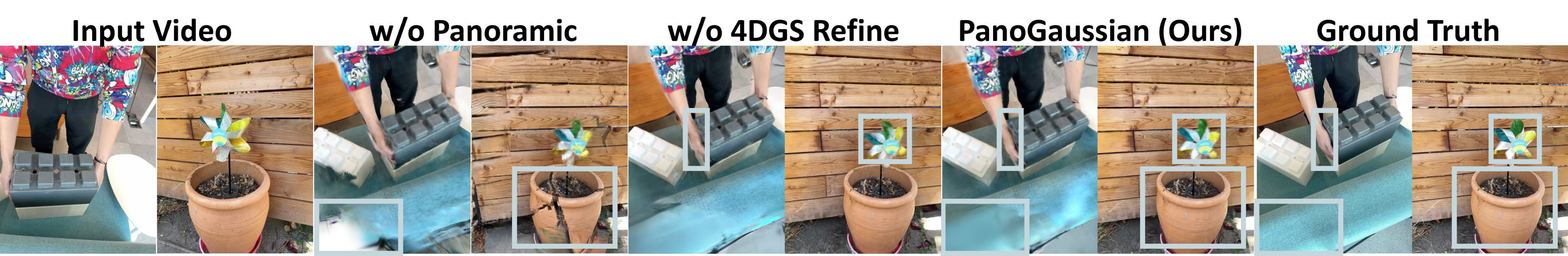}
\vspace{\figvspace}
\caption{\textbf{Panoramic design is vital for unseen-region consistency.}
Planar expansion with the same 4DGS backbone fails in unseen regions, while panoramic coordinates recover coherent geometry even without full Gaussian refinement.}
\label{fig:ablation_4dgs}
\vspace{\figvspace}
\end{figure}

\subsection{Ablation Studies}

\subsubsection{Necessity of Panoramic Representation.} 
To validate the necessity of our panoramic design, we disentangle panoramic expansion from 4DGS refinement on DyCheck~\cite{li2023dynibar}.
We compare planar expansion with the same 4DGS backbone, panoramic expansion without Gaussian refinement, and our full model in~\cref{tab:abl_comp} and~\cref{fig:ablation_4dgs}.
As shown in~\cref{tab:abl_comp}, planar expansion with the identical backbone substantially underperforms our full model on uPSNR, confirming that panoramic design is the prerequisite for unseen-region reconstruction.
Panoramic expansion alone already recovers most of this gap, while 4DGS refinement further improves overall visual quality in both seen and unseen regions.
As shown in~\cref{fig:ablation_4dgs}, the planar baseline fails to capture holistic scene structure, leading to severe geometric distortion and multi-view inconsistency when resolving large occlusions.
Our panoramic representation ensures physically coherent unconstrained 4D scene exploration.

\subsubsection{Panoramic Trajectories Training.} 
Training in panoramic trajectories provides limited quantitative gains in DyCheck~\cite{li2023dynibar} in~\cref{tab:abl_comp}. However, it yields clear qualitative improvements under significant viewpoint changes. 
As shown in~\cref{fig:ablation}, optimization along the designed panoramic trajectories produces results more consistent with the original scene. 
This is mainly attributed to the consistency in training and inference of \ourmethod{}, which enables a more stable and coherent view exploration.

\subsubsection{Progressive Expansion.}
\begin{figure}[t] 
\centering
\includegraphics[width=\columnwidth]{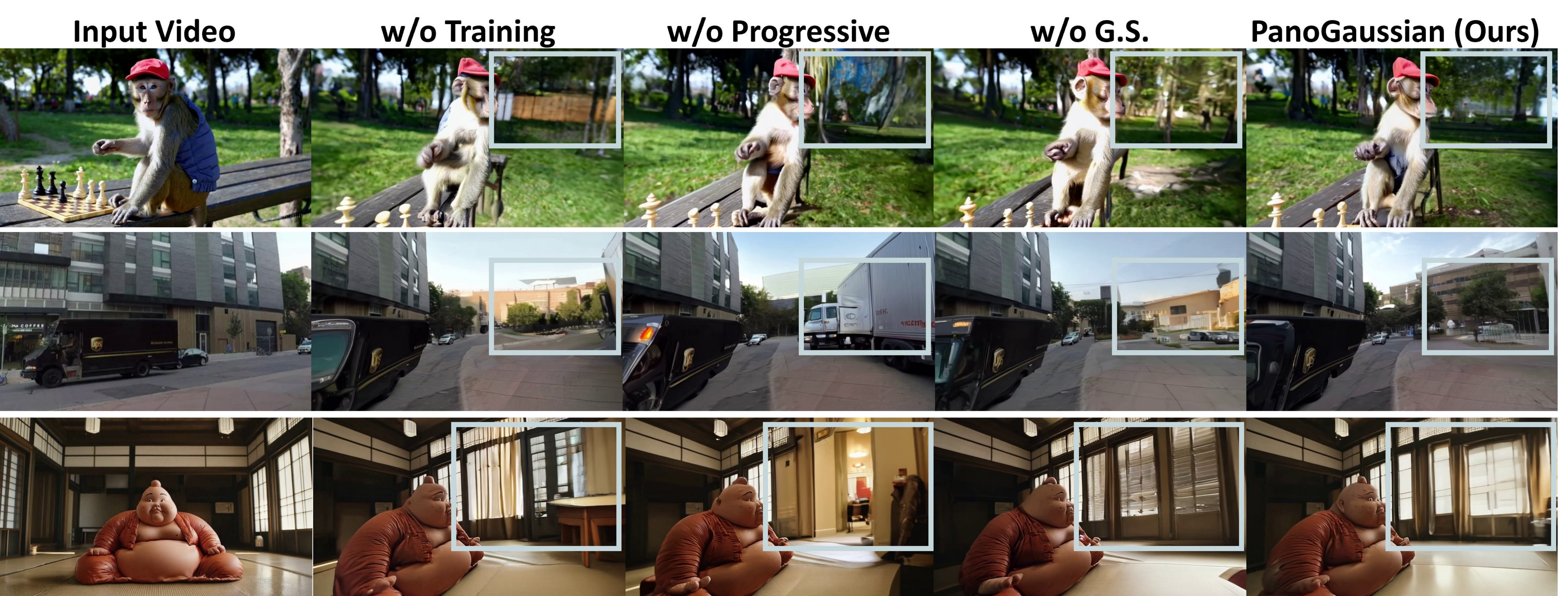}
\vspace{\figvspace}
\caption{
\textbf{Qualitative ablation study under extreme camera transitions.} 
The green boxed areas highlight regions with notable differences.
Training with panoramic trajectories and a progressive expansion strategy enables \ourmethod{} to infer unseen regions more consistently.
The unification with Gaussian Splatting representation further preserves physical realism and geometric fidelity.
}
\vspace{\figvspace}
\label{fig:ablation} 
\end{figure}

Our proposed progressive expansion strategy has significant contributions to explore and infer consistent scenes.
In~\cref{fig:ablation}, without this progressive strategy, direct one-step generation often relies on limited visual contexts, leading to fragmented or inconsistent results. 
In contrast, our progressive expansion gradually enlarges the explored region, allowing the model to reason over more comprehensive information.
This process effectively constrains generative priors and leads to smoother and more consistent reconstructions, especially under large camera transitions.
This strategy has also made a significant contribution to qualitative results in~\cref{tab:abl_comp}.
\setlength{\tabcolsep}{3pt} 
\begin{table}[t]
\centering
\caption{\textbf{Quantitative ablation on different components of the system.}
Best, second-best, and third-best results are highlighted in \colorbox{first}{red}, \colorbox{second}{orange}, and \colorbox{third}{yellow}, respectively.
Our proposed Panoramic representation and Gaussian Splatting representation yields the most significant quantitative improvement. The progressive expansion strategy further contributes to stable and coherent reconstruction.
}
\label{tab:abl_comp}
\vspace{\figvspace}
\resizebox{\columnwidth}{!}{%
\begin{tabular}{lcccccc}
\toprule
Method & $\uparrow$ PSNR & $\uparrow$ SSIM & $\downarrow$ LPIPS & $\uparrow$ uPSNR & $\uparrow$ uSSIM & $\downarrow$ uLPIPS \\
\midrule
Planar with 4DGS  & 16.47 & 0.550 & 0.377 & 14.25 & 0.302 & 0.445 \\
Pano without 4DGS refinement & 17.12 & 0.555 & 0.358 & 15.08 & 0.421 & 0.418 \\
Without Panoramic Training & \cellcolor{second}{\underline{18.46}} & \cellcolor{second}{\underline{0.584}} & \cellcolor{second}{\underline{0.327}} & \cellcolor{second}{\underline{16.32}} & \cellcolor{second}{\underline{0.476}} & \cellcolor{second}{\underline{0.397}} \\
Without Progressive Expansion & \cellcolor{third}{17.51} & \cellcolor{third}{0.576} & \cellcolor{third}{0.351} & \cellcolor{third}{15.22} & \cellcolor{third}{0.446} & \cellcolor{third}{0.417} \\
Without Gaussian Splatting & 14.67 & 0.392 & 0.445 & 12.61 & 0.314 & 0.482 \\
\midrule
\ourmethod{} (Ours) & \cellcolor{first}{\textbf{18.71}} & \cellcolor{first}{\textbf{0.598}} & \cellcolor{first}{\textbf{0.323}} & \cellcolor{first}{\textbf{16.72}} & \cellcolor{first}{\textbf{0.495}} & \cellcolor{first}{\textbf{0.385}} \\
\bottomrule
\end{tabular}}
\vspace{\figvspace}
\end{table}

\subsubsection{Gaussian Splatting Representation.}
As shown in~\cref{tab:abl_comp}, the injection of the Gaussian Splatting representation significantly contributes to the improvement of qualitative results in the DyCheck IPhone dataset. 
The geometric priors from Dynamic Gaussian Splatting encourage faithful alignment with the original monocular video.
This geometric alignment significantly reduces structural distortions and unrealistic artifacts, which improves physical consistency with the original monocular videos. 
As shown in~\cref{fig:ablation}, a visual failure example without Gaussian Splatting clearly exhibits temporal inconsistency, while our full unification improves both visual fidelity and geometric integrity, offering the physical consistency of 4D scenes.

\subsubsection{Error Propagation Analysis.}
Our design structurally prevents cascading errors during progressive expansion.
At every expansion step, warping is performed from the \textit{original} observed video $V_{\text{src},\text{inf}}$ using fixed 4D GS depth and poses, rather than from previously hallucinated outputs.
The binary mask $M_e$ restricts both diffusion inpainting and refinement loss to newly generated regions only, preventing re-optimization of already settled areas.
As shown in~\cref{tab:expansion_stability}, PSNR and uPSNR remain stable across expansion steps, confirming that each step's quality is independently determined by the fixed 4D GS reconstruction rather than accumulated from prior hallucinations.
\begin{table}[t]
\centering
\caption{\textbf{Reconstruction quality across expansion steps on DyCheck~\cite{li2023dynibar}.}  It confirms that errors do not accumulate during progressive expansion.}
\label{tab:expansion_stability}
\vspace{\figvspace}
\resizebox{\columnwidth}{!}{%
\renewcommand\arraystretch{1.05}%
\begin{tabular}{ccccc}
\toprule
Expansion Step & 6 & 12 & 18 & 24 \\
\midrule
PSNR / uPSNR (dB) & 18.71 / 16.72 & 18.72 / 16.74 & 18.69 / 16.68 & 18.68 / 16.66 \\
\bottomrule
\end{tabular}}
\vspace{\figvspace}
\end{table}

\section{Conclusion}
\label{sec:conclusion}
We propose \ourmethod{}, a unified Panoramic-Gaussian representation for the synthesis of geometry-aware 4D scenes. The panoramic representation, equipped with a progressive expansion strategy, enables a comprehensive exploration of the whole scene. The Gaussian  representation ensures high geometric fidelity by explicitly modeling spatial structures. Extensive experiments demonstrate that \ourmethod{} achieves consistent 4D scene synthesis, maintaining strong geometric accuracy even under large viewpoint variations. We expect that this unified representation may facilitate future work on physically grounded and controllable 4D content generation.

\subsubsection{Limitations.}
\ourmethod{} is most effective on inward-facing, object-centric scenes with sufficient inter-view overlap to anchor geometry.
On outward-facing captures, distant content provides weaker multi-view constraints and quality becomes less reliable.
The pipeline further relies on a pretrained video diffusion model and per-scene 4DGS optimization (${\sim}$2.0\,h per scene), which limits real-time deployment and may inherit artifacts from the generative prior.

\section*{\ackname}
This work was supported in part by New Generation Artificial Intelligence-National Science and Technology Major Project (2025ZD0123004), Ningbo grant (2025Z038) and National Natural Science Foundation of China (Grant \\ No.~62376060).

%
%
\bibliographystyle{splncs04}
\bibliography{main}

\ifarxiv
\clearpage
\appendix

\section{Future Works}

To our knowledge, \ourmethod{} is the first 4D reconstruction attempt that extends view interpolation to synthesis with unseen regions. While it achieves consistent and geometry-aware 4D  synthesis, several directions remain for future works.

\ourmethod{} currently adopts MoSca~\cite{lei2024mosca} as Gaussian representation, which requires per-scene optimization. This restricts generalization and prevents real-time 4D synthesis.
Another promising direction is to develop a feed-forward version of \ourmethod{} based on recent feed-forward Gaussian representations~\cite{liang2024feed, lin2025movies, xu20254dgt}, once open-source codes become available. 

Beyond synthesizing the entire scene with \ourmethod{}, an additional direction is to integrate editing and re-animation techniques for interactive scene manipulation. This capability would enable goal-driven navigation and immersive exploration of the generated scene for Virtual Reality and Augmented Reality. They may also support Multi-Agent Decision Making for embodied AI, where agents can plan and act within a controllable 4D scene from our framework.

Our \ourmethod{} representation is highly compatible with different Gaussian representations and video generative priors. Consequently, we expect that advances in either component will naturally improve perceptual quality.

\section{Outward-Facing Scene Evaluation}
\label{sec:suppl_outward}

\ourmethod{} is primarily evaluated on inward-facing, object-centric captures in the main paper.
To examine behavior under weaker geometric constraints, we additionally test outward-facing outdoor sequences where distant content lacks reliable cross-view 3D overlap.
As shown in~\cref{fig:suppl_outward}, nearby structures remain visually coherent, while distant regions degrade gracefully rather than collapsing.

\begin{figure}[t]
\centering
\includegraphics[width=\linewidth]{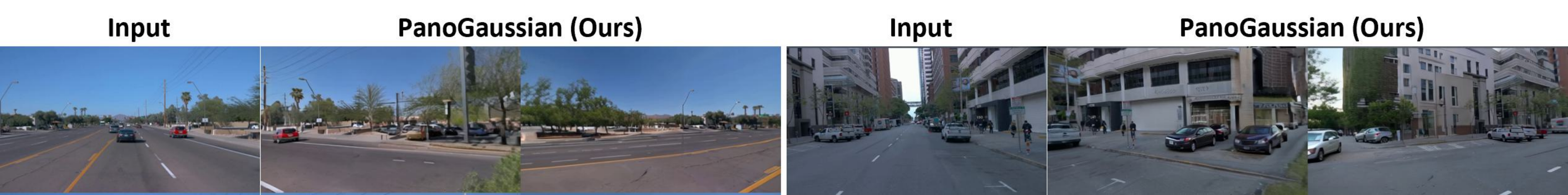}
\vspace{\figvspace}
\caption{\textbf{Results on outward-facing scenes.}
\ourmethod{} maintains visual consistency on nearby outdoor structures; distant regions degrade gracefully rather than collapsing.}
\label{fig:suppl_outward}
\vspace{\figvspace}
\end{figure}

\section{Quantitative Results on Kubric-4D}
\label{sec:suppl_kubric}

Following DyCheck, we report covisible-region metrics (mPSNR, mSSIM, mLPIPS) together with unseen-region metrics (uPSNR, uSSIM, uLPIPS) on Kubric-4D~\cite{vanhoorick2024gcd}, where large camera displacements create genuinely unobserved areas.
As summarized in~\cref{tab:suppl_kubric}, existing baselines fail to produce meaningful reconstruction in unseen regions, whereas \ourmethod{} maintains strong performance in both covisible and unseen areas.

\begin{table}[t]
\centering
\caption{\textbf{Comparison on Kubric-4D~\cite{vanhoorick2024gcd}} with large camera displacement.
Existing baselines fail in unseen regions, while \ourmethod{} maintains structural and visual quality.}
\label{tab:suppl_kubric}
\vspace{\figvspace}
\resizebox{\columnwidth}{!}{%
\renewcommand\arraystretch{1.05}%
\begin{tabular}{lcccccc}
\toprule
 & \multicolumn{3}{c}{Covisible Region} & \multicolumn{3}{c}{Unseen Region} \\
\cmidrule(lr){2-4} \cmidrule(lr){5-7}
Method & mPSNR$\uparrow$ & mSSIM$\uparrow$ & mLPIPS$\downarrow$ & uPSNR$\uparrow$ & uSSIM$\uparrow$ & uLPIPS$\downarrow$ \\
\midrule
DynPoint~\cite{zhou2024dynpoint} & 23.93 & 0.837 & 0.143 & 12.13 & 0.364 & 0.291 \\
MoSca~\cite{lei2024mosca} & 23.92 & 0.840 & 0.134 & 10.86 & 0.345 & 0.296 \\
TrajectoryCrafter~\cite{yu2025trajectorycrafter} & 16.76 & 0.396 & 0.367 & 15.53 & 0.389 & 0.205 \\
\textbf{\ourmethod{} (Ours)} & \textbf{26.15} & \textbf{0.879} & \textbf{0.101} & \textbf{17.19} & \textbf{0.454} & \textbf{0.157} \\
\bottomrule
\end{tabular}}
\vspace{\figvspace}
\end{table}

\section{Panoramic Projection Details}
\label{sec:suppl_pano_proj}

For each point $p_{t}^{j}=(x_{t}^{j}, y_{t}^{j}, z_{t}^{j}) \in P_{t}$, we first transform it into the coordinate system centered at the panorama origin $P^{c}$:
\begin{equation}
\tilde{p}_{t}^{j} = 
\begin{bmatrix}
\tilde{x}_{t}^{j} \\ \tilde{y}_{t}^{j} \\ \tilde{z}_{t}^{j}
\end{bmatrix}
=
R_{\text{align}}^{\top}
\begin{bmatrix}
x_{t}^{j} - x^c \\[3pt]
y_{t}^{j} - y^c \\[3pt]
z_{t}^{j} - z^c
\end{bmatrix}
   \label{eq:transformation}
\end{equation}
where $R_{\text{align}}$ aligns the first camera view with the positive $x$-axis of the panorama.
We then compute spherical coordinates:
\begin{equation}
   \phi_t^j = \arctan2(\tilde{y}_{t}^{j},\, \tilde{x}_{t}^{j}), \quad
   \theta_t^j = \arctan2(\tilde{z}_{t}^{j},\, 
      \sqrt{(\tilde{x}_{t}^{j})^2 + (\tilde{y}_{t}^{j})^2}).
\label{eq:pano1}
\end{equation}
Finally, each 3D point is projected onto the 2D panorama:
\begin{equation}
   u_t^j = \left(\frac{\phi_t^j}{2\pi} + 0.5\right) \times w_p,
   \quad
   v_t^j = \left(0.5 - \frac{\theta_t^j}{\pi}\right) \times h_p,
   \label{eq:pano2}
\end{equation}
where $w_p$ and $h_p$ denote the panoramic width and height.

\section{Full 4D Scene Synthesis}
We visualize full 4D scene synthesis with multi-view renderings and different timestamps in~\cref{fig:4d}. \ourmethod{} achieves temporally coherent and geometrically consistent 4D scene synthesis. The synthesized 4D scenes remain stable across viewpoint changes.

\begin{figure*}[ht] 
\centering
\includegraphics[width=\linewidth]{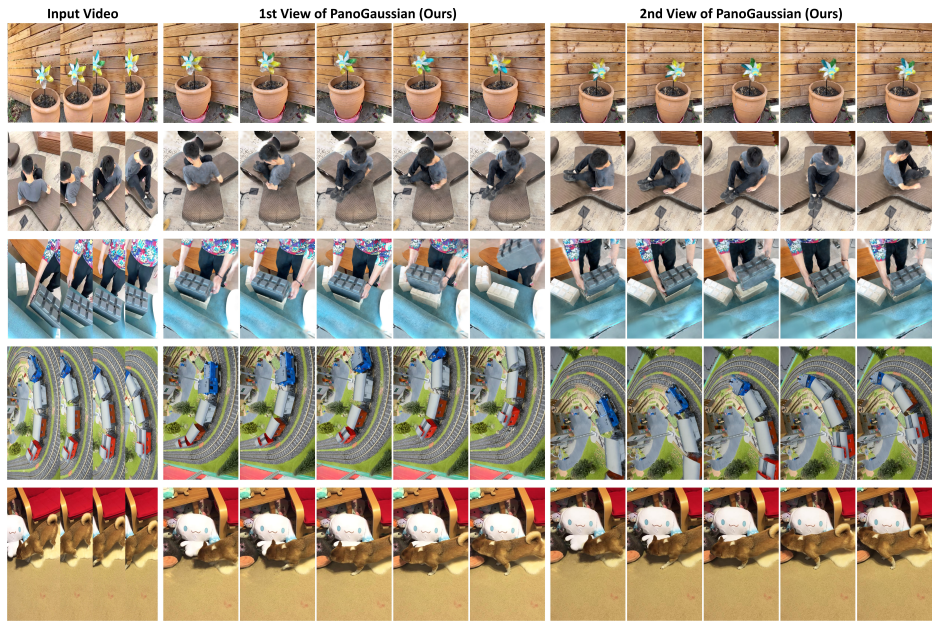}
\vspace{\figvspace}
\caption{
\textbf{Full 4D scene synthesis with \ourmethod{}.}
\ourmethod{} produces complete and geometry-consistent 4D reconstructions across multiple viewpoints and timestamps.
}
\vspace{\figvspace}
\label{fig:4d} 
\end{figure*}

\fi

\end{document}